\documentclass[10pt,twocolumn,letterpaper]{article}

\usepackage{cvpr}              % To produce the 
\usepackage[accsupp]{axessibility}

\usepackage[dvipsnames]{xcolor}
\definecolor{cvprblue}{rgb}{0.21,0.49,0.74}
\usepackage[pagebackref,breaklinks,colorlinks,citecolor=cvprblue]{hyperref}

\def\paperID{15576} % *** Enter the Paper ID here
\def\confName{CVPR}
\def\confYear{2024}

\usepackage{algorithm}
\usepackage{algorithmic}
\usepackage{amsmath}
\usepackage{amssymb}
\usepackage{amsfonts}
\usepackage{amsthm}
\newtheorem{theorem}{Theorem}

\newtheorem{definition}{Definition}
\usepackage{multirow}

\usepackage{hhline}

\title{Towards General Robustness Verification of MaxPool-based Convolutional Neural Networks via Tightening Linear Approximation}

\author{
    Yuan Xiao$^{1}$, Shiqing Ma$^{2}$, Juan Zhai$^{2}$,  Chunrong Fang$^{1}$\thanks{Chunrong Fang and Zhenyu Chen are the corresponding authors.}, Jinyuan Jia$^{3}$, Zhenyu Chen$^{1, 4*}$\\
    $^{1}$ State Key Laboratory for Novel Software Technology, Nanjing University, China \\ %\quad
    $^{2}$University of Massachusetts Amherst, United States \quad 
    $^{3}$ Pennsylvania State University, United States\\
       $^{4}$ Shenzhen Research Institute, Nanjing University, China
}

\begin{document}

\maketitle

\begin{abstract}
    The robustness of convolutional neural networks (CNNs) is vital to modern AI-driven systems. It can be quantified by formal verification by providing a certified lower bound, within which any perturbation does not alter the original input's classification result. It is challenging due to nonlinear components, such as MaxPool. At present, many verification methods are sound but risk losing some precision to enhance efficiency and scalability, and thus, a certified lower bound is a crucial criterion for evaluating the performance of verification tools. In this paper, we present \textbf{MaxLin}, a robustness verifier for \textbf{Max}pool-based CNNs with tight \textbf{Lin}ear approximation. By tightening the linear approximation of the MaxPool function, we can certify larger certified lower bounds of CNNs. We evaluate MaxLin with open-sourced benchmarks, including LeNet and networks trained on the MNIST, CIFAR-10, and Tiny ImageNet datasets. The results show that MaxLin outperforms state-of-the-art tools with up to 110.60\% improvement regarding the certified lower bound and 5.13 $\times$ speedup for the same neural networks. Our code is available at \href{https://github.com/xiaoyuanpigo/maxlin}{https://github.com/xiaoyuanpigo/maxlin}.
    
    \end{abstract}

\section{Introduction}
Convolutional neural networks (CNNs) have achieved remarkable success in various applications, such as speech recognition~\cite{xiong2016achieving} and image classification~\cite{nath2014survey}.
However, accompanied by outstanding effectiveness, neural networks are often vulnerable to environmental perturbation and adversarial attacks~\cite{moosavi2017universal,szegedy2013intriguing}.
Such fragility will lead to disastrous consequences in safety-critical domains, e.g., self-driving~\cite{gopinath2018deepsafe} and face recognition~\cite{goswami2018unravelling}.
Therefore, a formal and deterministic robustness guarantee is indispensable before a network is deployed~\cite{balunovic2019certifying}.

The methodology of robustness verification can be divided into two categories: complete verifiers and incomplete verifiers.  
Complete methods~\cite{katz2017reluplex,katz2019marabou} can verify the robustness of piece-wise linear networks without losing any precision but fail to work on more complex network structures~\cite{li2020sok}.
Incomplete but sound verification~\cite{henriksen2020efficient,wu2021tightening,singh2019boosting,zhang2022provably} aims to scale to different types of CNNs. 
The major challenge of robustness verification of CNNs stems from their non-linear properties. 
Most incomplete verifiers~\cite{henriksen2020efficient,wu2021tightening,xu2020fast,wang2021beta,lyu2020fastened} focus on the ReLU- and Sigmoid-based networks whose activations are uni-variate functions and are simple to verify, ignoring multi-variate functions like MaxPool. 
Multi-variate function MaxPool is widely adopted in CNNs~\cite{liu2022convnet,he2016deep,xie2017aggregated} yet is far more complex to verify. 
Until recently, some attempts~\cite{boopathy2019cnn, singh2019abstract, lorenz2021robustness, xiao2022certifying} have been made to certify the robustness of MaxPool-based CNNs. 
Unfortunately, many of these verification frameworks~\cite{wang2021beta,xu2020fast,singh2019abstract} can only certify $l_\infty$ perturbation form. Furthermore, these existing methods are limited in terms of (1) efficiency: multi-neuron relaxation~\cite{muller2022prima} fail to scale to larger models due to long calculation time; (2) precision: single-neuron relaxation~\cite{boopathy2019cnn, singh2019abstract, lorenz2021robustness, xiao2022certifying} has loose certified lower bounds because of imprecise approximation.
\begin{figure*}[t] 
\centering
\includegraphics[width=1.0\textwidth]{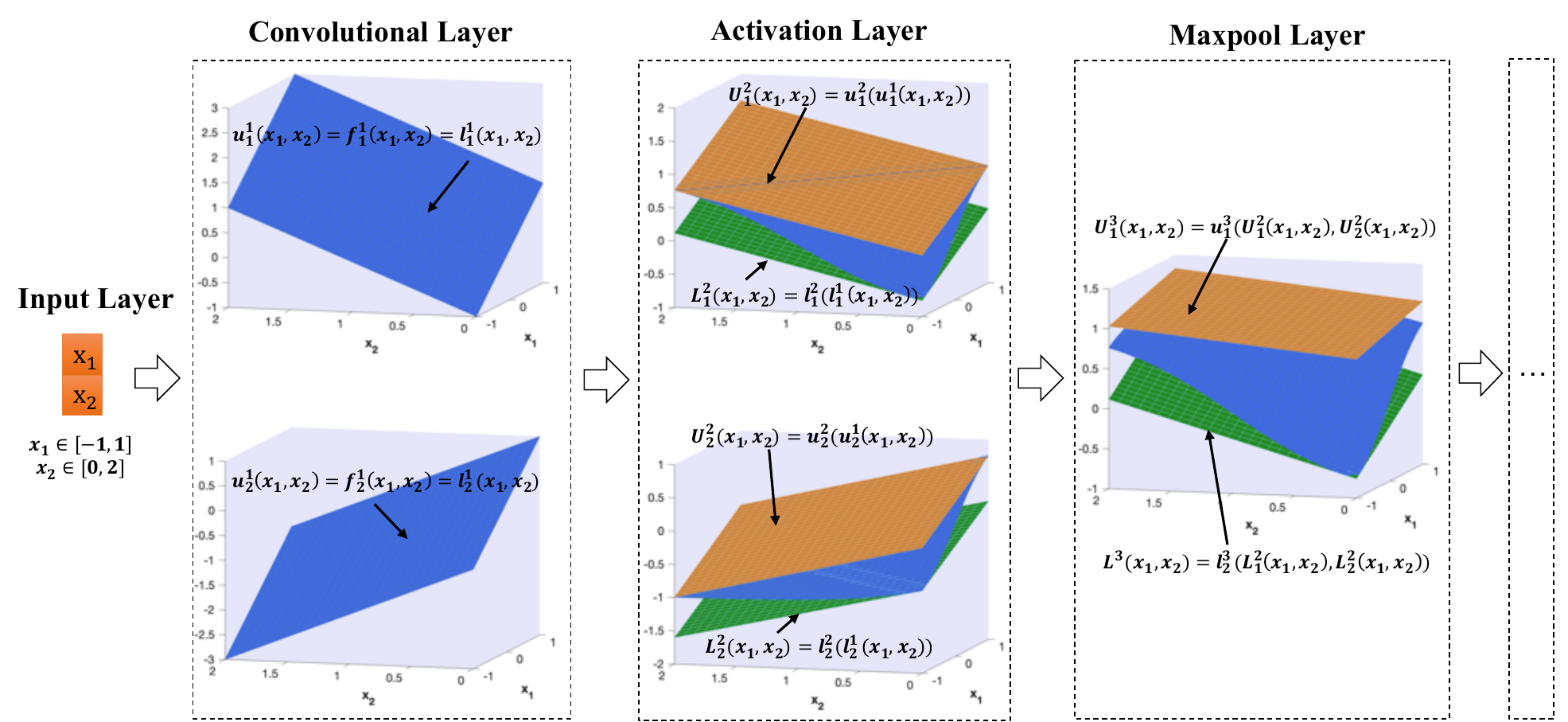} 
\caption{A toy example of MaxLin linear approximation. 
To simplify, the input size is two, and the perturbation radius is one.
$l^k_i(x^k_i)$ and $u_i^k(x^k_i)$ are the lower and upper linear bounds of the output of the $i$-th neuron($x^k_i$) in the $k$-th layer, respectively.
$L^k_i(x_1,x_2)$ and $U_i^k(x_1,x_2)$ are the global lower and upper linear bounds of the output of the $i$-th neuron in the $k$-th layer, respectively. 
The blue surface is the output of the current neuron, and the activation function here is the Tanh function.}\label{fig1}
\end{figure*}

To address the above challenges, in this work, we propose MaxLin, an efficient and tight verification framework for MaxPool-based networks via tightening linear approximation. 
Specifically, to tighten linear approximation, we minimize the maximum value of the upper linear bound and minimize the average precision loss of the lower linear bound of MaxPool.
We also prove that our proposed upper bound is block-wise tightest. Compared with existing neuron-wise tightness, our method acheives better certified results.
Further, based on single-neuron relaxation, MaxLin gives the linear bounds directly after choosing the first and second maximum values of the upper and lower bound of the MaxPool's input.
Thus, MaxLin has high computation efficiency. 
A simple example of MaxLin's computation process is shown in Figure~\ref{fig1}. Moreover, MaxLin easily integrates with state-of-the-art verifiers, e.g., CNN-Cert~\cite{boopathy2019cnn}, 3DCertify~\cite{lorenz2021robustness}, and $\alpha,\beta$-CROWN~\cite{xu2020fast,wang2021beta}. 
The integration allows MaxLin to certify different types of MaxPool-based networks (e.g., CNNs or PointNet) with various activation functions (e.g., Sigmoid, Artan, Tanh or ReLU) against \(l_1, l_2, l_\infty \)-norm perturbations.

We evaluate MaxLin with open-sourced benchmarks on the MNIST~\cite{lecun1998mnist}, CIFAR-10~\cite{krizhevsky2009learning}, and Tiny ImageNet~\cite{deng2009imagenet} datasets. 
The experiment results show that MaxLin outperforms the state-of-the-art techniques including CNN-Cert~\cite{boopathy2019cnn}, DeepPoly~\cite{singh2019abstract}, 3DCertify~\cite{lorenz2021robustness}, and Ti-Lin~\cite{xiao2022certifying} with up to 110.60\%, 62.17\%, 39.94\%, and 49.26\% improvement in terms of tightness, respectively. 
MaxLin has higher efficiency with up to 5.13$\times$ speedup than 3DCertify and comparable efficiency as CNN-Cert, DeepPoly, and Ti-Lin. 
Further, we compare MaxLin with branch and bound (BaB) methods, including $\alpha$,$\beta$-CROWN~\cite{xu2020fast,wang2021beta,zhang2022general-gcp}, ERAN\footnote{ERAN:~~\url{https://github.com/eth-sri/eran}} and MN-BaB~\cite{ferrari2022complete}, on ERAN benchmarks. 
The results show that MaxLin has much higher certified accuracy and less time cost across different perturbation ranges.

In summary, our work proposes an incomplete robustness verification technique, MaxLin, with tighter linear approximation and better efficiency, which works for various CNNs and \(l_p\)-norm perturbations.
By tightening linear approximation for MaxPool, our approach outperforms the state-of-the-art tools with up to 110.60\% improvement to the certified robustness bounds and up to 5.13\(\times\) speedup.

\section{Related Work}
We now introduce some  topics closely related  to robustness verification and then introduce other related robustness verification techniques.
\subsection{Adversarial Attacks and Defenses}

Many research studies~\cite{szegedy2013intriguing,goodfellow2014explaining,chen2017zoo,chen2020hopskipjumpattack,carlini2017towards,silva2020opportunities,wu2023adversarial}  show machine learning models are vulnerable to adversarial examples. 
Adversarial examples pose severe concerns for the deployment of machine learning models in security and safety-critical applications such as autonomous driving.
To defend against adversarial examples, many defenses~\cite{goodfellow2014explaining,madry2018towards,katz2017reluplex,katz2019marabou,cohen2019certified,henriksen2020efficient,wu2021tightening,xu2020fast,wang2021beta,lyu2020fastened} were proposed. Empirical defenses~\cite{papernot2016distillation,buckman2018thermometer} cannot provide a formal robustness guarantee and they are often broken by adaptive, unseen attacks~\cite{carlini2017adversarial,athalye2018obfuscated,ghiasi2020breaking}. Thus, we study certified defenses in this work. In particular, we focus on MaxPool-based convolutional neural networks which are widely used for image classification.
\subsection{Robustness Verification for MaxPool-based CNNs}
As MaxPool is hard to verify, only a few research on robustness verification takes MaxPool into consideration.  Recently, a survey on certified defense is proposed~\cite{meng2022adversarial}. Verification approaches are usually divided into two classes: complete verification and incomplete verification. 

As for complete verifiers, Marabou~\cite{katz2019marabou} extends Reluplex~\cite{katz2017reluplex} and proposes a precise SMT-based verification framework to verify arbitrary piece-wise linear network, including ReLU-based networks with MaxPool layers. However, 
this complete method cannot apply to other non-linear functions, such as Sigmoid and Tanh.  Recently, 
PRIMA~\cite{muller2022prima} proposes a general verification framework based on multi-neuron relaxation and can apply to MaxPool-based networks. Further, MN-BaB~\cite{ferrari2022complete} proposes a complete neural network verifier that builds on the tight multi-neuron constraints proposed in PRIMA.
However,  multi-neuron relaxation methods may contain an exponential number of linear constraints at the worst case~\cite{tjandraatmadja2020convex} and cannot verify large models in a feasible time(one day per input)~\cite{li2020sok}. 

To break the scalability barrier of the above work and accelerate the verification process, linear approximation based on single-neuron relaxation has been created.
CNN-Cert~\cite{boopathy2019cnn} proposes an efficient verification framework with non-trivial linear bounds for MaxPool. However, CNN-Cert is loose in terms of tightness and only applies to layered CNNs and ResNet. DeepPoly~\cite{singh2019abstract} proposes a versatile verification framework for different networks. However, it certifies very loose robustness bounds and certifies robustness only against $l_\infty$ perturbation form.
Recently, 3DCertify propose a novel verifier built atop DeepPoly and can certify the robustness of PointNet. 
3DCertify uses the Double Description method to tighten the linear approximation for MaxPool. However, its linear approximation is still loose and it is time-consuming.
Ti-Lin~\cite{xiao2022certifying} proposes the neuron-wise tightest linear bounds for MaxPool by producing the smallest over-approximation zone. 
However, MaxPool often comes after ReLU, Sigmoid, or other non-linear layers, which pose a big challenge to tighten and thus, Ti-Lin is still loose in tightness.

\section{Preliminaries}
\label{section2}
This section introduces the minimal necessary background of our approach.

\subsection{MaxPool-based Neural Networks}

We focus on certifying the robustness of MaxPool-based networks for classification tasks. Our methods can refine the abstraction of the MaxPool function in arbitrary networks. For simplicity, we formally use $F:\mathbb{R}^{n_0}\to\mathbb{R}^{n_K}$ to represent a neural network classifier with (K+1) layers and 
 $F=f^K\circ f^{K-1}\circ \cdots f^{2}\circ f^1$.
 Here $f^1: \mathbb{R}^{n_0}\to \mathbb{R}^{n_1},\cdots, f^K: \mathbb{R}^{n_{K-1}}\to \mathbb{R}^{n_K}$.
The symbol $f^i, i=1,\cdots, K$ could be an affine, activation, fully connected, or MaxPool function.
In this work, the non-linear block in neural architectures could be activation or activation+MaxPool.  The MaxPool function is defined as follows.
$$MaxPool(x_{i_1},\cdots,x_{i_n})=max\{x_{i_1},\cdots,x_{i_n}\}$$
where $i_1, \cdots, i_n$ are the indexes of the input that will be pooled associated with the $i$-th output of the current layer. 

As for other notations used in our approach, $n_{k}$   represents the number of neurons in the $k$-th layer and $[K]$ represents the set $\{1,\cdots,K\}$.   $F_{j}^k(\boldsymbol{x}):\mathbb{R}^{n_0}\to \mathbb{R}$ to denote the $j$-th output of  the $k$-th layer and $\boldsymbol{x^{k-1}}$ to denote the input of the $k$-th layer.

\subsection{Robustness Verification For Neural Networks}

Robustness verification aims to find the minimal adversarial attack range. In other words, robustness verification can give the largest certified robustness bound, within which there exist no adversarial examples around the original input. 
Such the maximum absolute safe radius is defined as local robustness bound $\epsilon_r$,  which are the formal robustness guarantees provided by complete  verifiers.

Define $\boldsymbol{x_0}$ be an input data point.
 Let $\mathbb{B}_p(\boldsymbol{x_0},\epsilon)$ denotes $\boldsymbol{x_0}$ perturbed within an $l_p$-normed ball with radius $\epsilon$, that is $\mathbb{B}_p(\boldsymbol{x_0},\epsilon)=\{\boldsymbol{x}|\Vert \boldsymbol{x}-\boldsymbol{x_0}\Vert_p \leq \epsilon\}$. We focus on $l_1, l_2$, and $l_\infty$ adversary, i.e. $p=1, 2, \infty$. Let $t$ denote the true label of $\boldsymbol{x_0}$.
 Then local robustness bound is defined as follows.
\begin{definition}[Local robustness bound]
\label{local}
$F$ is a neural network and $\epsilon_r\geq0$. $\epsilon_r$ is called as the local robustness bound of the input $\boldsymbol{x_0}$ in the neural network $F$ if
 $(\mathop{\arg\max}\limits_{i} F_i(\boldsymbol{x})=t,\forall \boldsymbol{x}\in \mathbb{B}_p(\boldsymbol{x_0},\epsilon_r))$
$\wedge$ $(\forall \delta>0,\exists \boldsymbol{x_a}\in \mathbb{B}_p(\boldsymbol{x_0},\epsilon+\delta)s.t. \mathop{\arg\max}\limits_{i}F_i(\boldsymbol{x_a})\neq t)$. 
\end{definition}

It is of vital importance to certify local robustness bound for networks. However, it is an NP-complete problem for the simple ReLU-based fully-connected networks~\cite{katz2017reluplex} and computationally expensive with the worse case of exponential time complexity~\cite{meng2022adversarial}. Therefore, it is practical to lose some precision to certify a lower bound than $\epsilon_r$, which is provided by incomplete verifiers.
\begin{definition}[Certified lower bound]
$F$ is a neural network and  $\epsilon_{l}\geq0$. $\epsilon_{l}$ is called as a certified lower bound of the input $\boldsymbol{x_0}$  in the  neural network $F$ if $(\epsilon_{l}<\epsilon_r)$ $\wedge$ $(\mathop{\arg\max}\limits_{i} F_i(\boldsymbol{x})=t,\forall  \boldsymbol{x}\in \mathbb{B}_p(\boldsymbol{x_0},\epsilon_{l}))$.

\end{definition}

Because incomplete verifier risks precision loss to gain scalability and efficiency,   the value of $\epsilon_{l}$ becomes a key criterion to evaluate the tightness of robustness verification methods and is used as the metric for tightness in our approach. 
 
\subsection{Linear Approximation}

Define $\boldsymbol{l^{k-1}}, \boldsymbol{u^{k-1}}$ are the lower and upper bound of the input of the $k$-th layer, that is,  $\boldsymbol{x^{k-1}}\in[\boldsymbol{l^{k-1}},\boldsymbol{u^{k-1}}]$.
The essence of linear approximation technique is giving linear bounds to every layer, that is $\forall k\in[K]$,  $l^k(\boldsymbol{x^{k-1}})\leq f^k(\boldsymbol{x^{k-1}})\leq u^k(\boldsymbol{x^{k-1}}),\forall \boldsymbol{x^{k-1}}\in [\boldsymbol{l^{k-1}},\boldsymbol{u^{k-1}}]$. 
\begin{definition}[Upper/Lower linear bounds]
Let $s^i$ be the input associated with the the $i$-th neuron ouput and
$f^k_i(\boldsymbol{s^{i,k-1}})$ be the function of the $i$-th neuron in the  $k$-th layer of neural network $F$. With $\boldsymbol{x^{k-1}}\in[\boldsymbol{l^{k-1}},\boldsymbol{u^{k-1}}]\subset \mathbb{R}^{n_{k-1}},$ 
if $\boldsymbol{s^{i,k-1}}\subset \mathbb{R}^{n}$ and there exists $\boldsymbol{a^k_u},\boldsymbol{a^k_l}\in\mathbb{R}^{n}$ and $\boldsymbol{b^k_u},\boldsymbol{b^k_l}\in \mathbb{R}$ such that $\forall \boldsymbol{s^{i,k-1}}\subset \boldsymbol{x^{k-1}}\in[\boldsymbol{\boldsymbol{l^{k-1}}},\boldsymbol{u^{k-1}}]$,
$$u^k_i(\boldsymbol{s^{i,k-1}})=\boldsymbol{a^k_{u}}\boldsymbol{s^{i,k-1}}+\boldsymbol{b^k_{u}}, l^k_i(\boldsymbol{x^{k-1}})=\boldsymbol{a^k_{l}}\boldsymbol{s^{i,k-1}}+\boldsymbol{b^k_{l}}$$
$$l^k_i(\boldsymbol{s^{i,k-1}})\leq f^k_i(\boldsymbol{s^{i,k-1}})\leq u^k_i(\boldsymbol{s^{i,k-1}})$$
then, $u^k_i(\boldsymbol{s^{i,k-1}})$ and $l^k_i(\boldsymbol{s^{i,k-1}})$ are called  upper and lower linear  bounds of $f^k_i(\boldsymbol{s^{i,k-1}})$, respectively.
\end{definition}

It is worth mentioning that $n$ is determined by the type of $f^k_i(\boldsymbol{s^{i,k-1}})$. When $f^k_i(\boldsymbol{s^{i,k-1}})$ is a univariate function(such as  ReLU, Sigmoid, Tanh, or Arctan), $n=1$. When $f^k_i(\boldsymbol{s^{i,k-1}})$ is a multivariate function, $n$ is equal to the dimension of $\boldsymbol{s^{i,k-1}}$. For example, when $f^k_i(\boldsymbol{s^{i,k-1}})$ is MaxPool, $n$ is equal to the size of the input to be pooled; 
When the $k$-th layer is a convolutional layer, $n$ corresponds to the size of the weight filter, and the linear constraints are $$u^k(\boldsymbol{s^{i,k-1}})= \boldsymbol{w}*\boldsymbol{s^{i,k-1}}+\boldsymbol{b},l^k(\boldsymbol{s^{i,k-1}})= \boldsymbol{w}*\boldsymbol{s^{i,k-1}}+\boldsymbol{b}$$
where $*$ is the convolution operation. $\boldsymbol{w}$ and $\boldsymbol{b}$ are the filter's weights and biases, respectively.
When the $k$-th layer is a fully-connected layer, $n=n_{k-1}$ and the linear constraints are $$u^k(\boldsymbol{x^{k-1}})= \boldsymbol{w}\boldsymbol{s^{i,k-1}}+\boldsymbol{b},l^k(\boldsymbol{s^{i,k-1}})= \boldsymbol{w}\boldsymbol{s^{i,k-1}}+\boldsymbol{b}$$
where $\boldsymbol{w}$ and $\boldsymbol{b}$ are the weights and biases  assosicated with the $i$-th ouput neuron in the fully-connected layer, respectively.

After giving linear constraints to the predecessor layers, we can compute the global linear bounds of the current layer, which is represented as:
$$L^k(\boldsymbol{x^{0}}):=\boldsymbol{A^k_l} \boldsymbol{x^{0}}+\boldsymbol{B_l^k},  U^k(\boldsymbol{x^{0}}):=\boldsymbol{A^k_u} \boldsymbol{x^{0}}+\boldsymbol{B_u^k}$$
where $ L^k(\boldsymbol{x^{0}}) \leq F^k(\boldsymbol{x^{0}})\leq U^k(\boldsymbol{x^{0}}), \forall \boldsymbol{x^0}\in \mathbb{B}_p(\boldsymbol{x_0},\epsilon)$.
The whole procedure is a layer-by-layer process from the first hidden layer to the last output layer, and we can compute a certified lower bound after we get the global linear bounds of the output layer.

\section{MaxLin: A Robustness Verifier for MaxPool-based CNNs} 
In this section, we present MaxLin, a  tight and efficient robustness verifier for MaxPool-based networks.

\subsection{Tightening Linear Approximation for MaxPool}

In this subsection, we propose our MaxPool linear bounds. We use $f(x_{1},\cdots,x_{n})=max\{x_{1},\cdots,x_{n}\}$ to represent the MaxPool function without loss of generality.

\begin{theorem}
\label{theoremMaxPool}
Given $f(x_1,\cdots,x_n)=max\{x_1,\cdots,x_n\}$, $x_i\in[l_i,u_i]$, we select the first and the second maximum values of the set $\{u_i|i=1,\cdots,n\}$ and assume their indexs are $i,j$, respectively. We use $l_{max}$ to denote the maximum value of the set $\{l_i|i=1,\cdots,n\}$. Define $\boldsymbol{m}=(m_1,\cdots,m_n)=(\frac{u_1+l_1}{2},\cdots,\frac{u_n+l_n}{2})  \in\mathbb{R}^n$. Then, the linear bounds of the MaxPool function are:

\textbf{Upper linear bound:}

$u(x_1,\cdots,x_n):=\sum_ia_i(x_i-l_i)+b$. 
Specifically, there are two different cases:

Case 1:
If $(l_i=l_{max})\wedge(l_i\geq u_j)$, $a_i=1;b=l_i;a_k=0,\forall k\neq i$.

Case 2:
Otherwise, $a_i=\frac{u_i-u_j}{u_i-l_i};b=u_j;a_k=0,\forall k\neq i$.

\textbf{Lower linear bound:}

$l(x_1,\cdots,x_n)=x_j,j=\mathop{\arg\max}\limits_{i}(m_i)$.
\end{theorem}

\subsection{Block-wise Tightest Property}

Existing methods~\cite{singh2019abstract,boopathy2019cnn,ko2019popqorn,henriksen2020efficient,xiao2022certifying} give the neuron-wise tightest linear bounds, producing the smallest the over-approximation zone for the ReLU, Sigmoid, Sigmoid($x$)Tanh($y$),$x$$\cdot$Sigmoid($y$) and MaxPool functions, respectively.
This notion ignores the interleavings of neurons and leads to non-optimal results.
In this paper, we introduce the notion of block-wise tightest, that is, the volume of the over-approximation zone between the global linear bounds of the ReLU+MaxPool block is the minimum.
This notion considers the interleavings of neurons, and the achieved results will be superior to existing neuron-wise tightest.
Without loss of generality, we assume Activation is at the $k$-th layer, and we use $U^{k+1}_b(\cdot)$ and $L^{k+1}_b(\cdot)$ to denote the global upper and lower linear bounds of the Activation+MaxPool block, respectively. 
Then, we define  the notion of block-wise tightest as follows:

\begin{definition}[Block-wise Tightest]
    The global linear bounds of the Activation+MaxPool block are $U^{k+1}_b(\boldsymbol{x^k})$ and $L^{k+1}_b(\boldsymbol{x^k})$, respectively. Then, we define $U^{k+1}_b(\boldsymbol{x^k})$ and $L^{k+1}_b(\boldsymbol{x^k})$ is the block-wise tightest if and only if 
    $\iint_{\boldsymbol{x^{k-1}}\in[\boldsymbol{l^{k-1},\boldsymbol{u^{k-1}]}}}(U^{k+1}_b(\boldsymbol{x^{k-1}})-L^{k+1}_b(\boldsymbol{x^{k-1}}))d\boldsymbol{x^{k-1}}$ reach the minimum.
\end{definition}

Furthermore, if the non-linear block is ReLU+MaxPool and the abstraction for ReLU is not precise and instead uses the neuron-wise tightest upper linear bound, then  MaxLin has the provably block-wise tightest upper linear bound. 
\begin{theorem}
    \label{theorem provable tightest}
    If  the preceding layer of the MaxPool function is ReLU with $u(x)=\frac{u}{u-l}(x-l)$ as the upper linear bound~\cite{boopathy2019cnn,singh2019abstract,xu2020fast}, the upper linear bound in Theorem \ref{theoremMaxPool} is the block-wise tightest.
\end{theorem}

We put the proofs of Theorem \ref{theoremMaxPool} and Theorem \ref{theorem provable tightest} in the supplementary material.

\begin{figure} 
\centering
\includegraphics[width=1.03\columnwidth]{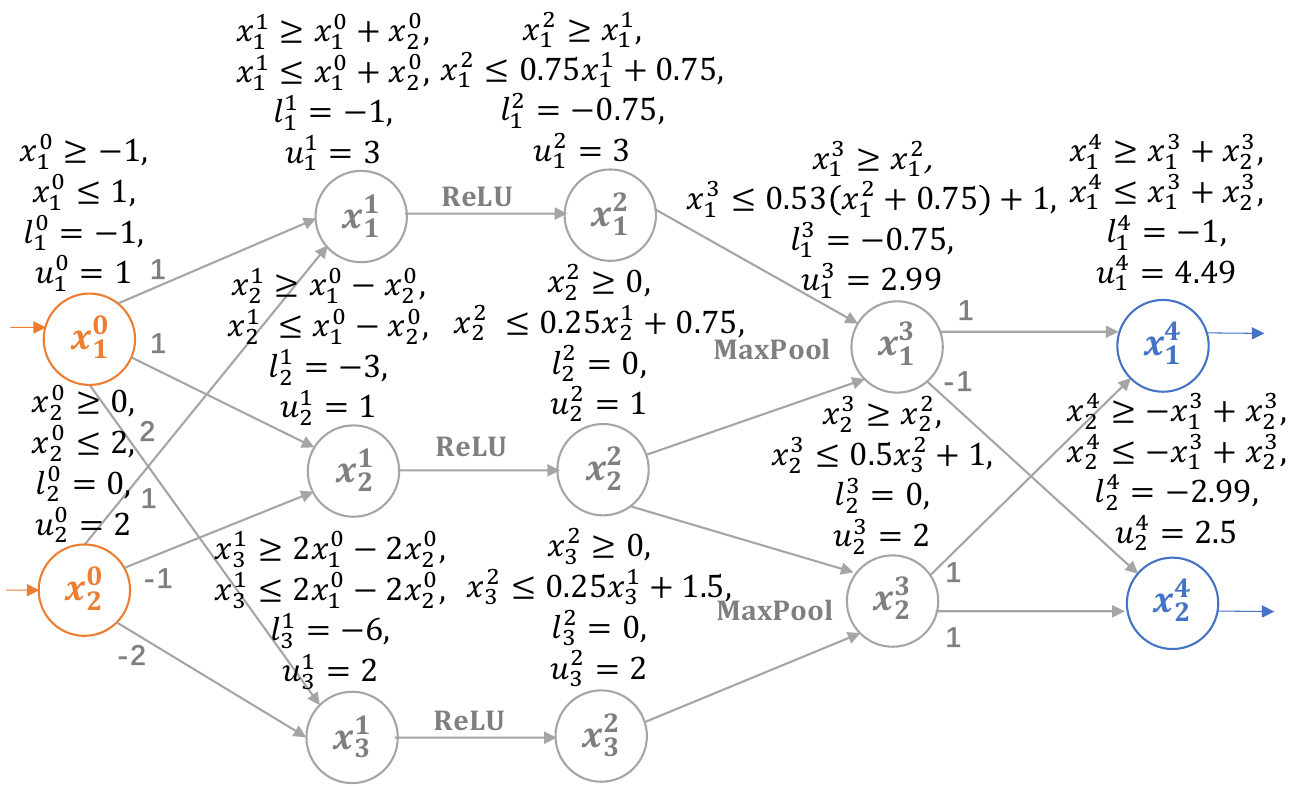} 
\caption{A toy example of how MaxLin computing global bounds $\boldsymbol{l^K}$ and $\boldsymbol{u^K}$ against $l_\infty$ adversary. The first, second, third, and fourth hidden layers are the affine, ReLU, MaxPool, and affine functions, respectively.}
\label{fig2}
\end{figure}

\subsection{Computing Certified Lower Bounds}
\label{section-compute}

The whole process of computing certified lower bounds can be divided into two parts: (i) computing the global upper and lower bounds $\boldsymbol{l^K},\boldsymbol{u^K}$ of the  network output $F^K(\boldsymbol{x})$ and (ii) searching the maximal certified lower bound.

\subsubsection{Computing the global upper and lower bounds $\boldsymbol{l^K},\boldsymbol{u^K}$ of the  network output $F^K(\boldsymbol{x})$}

Given a certain perturbation range $\epsilon$ and an original input $\boldsymbol{x_0}$, MaxLin can tightly compute the global upper and lower bounds $\boldsymbol{l^K},\boldsymbol{u^K}$ of the network output $F^K(\boldsymbol{x})$ to check whether  $\epsilon$  is a certified safe perturbation radius or not.

This process starts at a basic step, that is, we give a pair of linear bounds with the input range $\mathbb{B}_p(\boldsymbol{x_0},\epsilon)$ of $f^1(\boldsymbol{x})$, and then with $k=1$, we compute $\boldsymbol{l^1}$, $\boldsymbol{u^1}$ based on Equation (1) which are deduced~\cite{wu2021tightening} by Holder's inequality.
\begin{equation}
\begin{aligned}
    &F^k(\boldsymbol{x})\leq \epsilon \Vert \boldsymbol{A^k_u}\Vert_q +\boldsymbol{A^k_u} \boldsymbol{x_0}+\boldsymbol{B^k_u}, \\
    & F^k(\boldsymbol{x})\geq -\epsilon \Vert \boldsymbol{A^k_l}\Vert_q +\boldsymbol{A^k_l} \boldsymbol{x_0}+\boldsymbol{B^k_l}
\end{aligned}
\end{equation}
where $\Vert\cdot \Vert_q$ is $l_q$ norm and  $\frac{1}{p}+\frac{1}{q}=1$. In our work, we focus on \(l_1, l_2, l_\infty \)-norm  adversary and thus,  $p=1, 2, \infty$.

In the second step, without loss of generality, we assume the current layer is the $k$-th layer. Given $\boldsymbol{l^{k-1}}, \boldsymbol{u^{k-1}}$, we give upper and lower  linear bounds $u^k(\boldsymbol{x}), l^k(\boldsymbol{x})$, respectively. Then, we use Equation (1) to attain $\boldsymbol{l^k}, \boldsymbol{u^k}$ by backsubstitution~\cite{singh2019abstract}, which we will illustrate in detail later.  $k$ in the second step can be all positive integers that are smaller than $K$. Repeating the second step from $k=2$ to $k=K$, we can get the value of $\boldsymbol{l^K}$ and $\boldsymbol{u^K}$. If $l^K_t\geq u^K_j, \forall j\neq t, j\in[n_K]$, $\epsilon$ is a certified safe perturbation radius. Otherwise, $\epsilon$ cannot be certified to be a safe perturbation radius.

\textbf{A toy example.} To better illustrate the process of backsubstitution, we give a  toy example of how we compute $\boldsymbol{l^K}$ and $\boldsymbol{u^K}$ of a five-layer fully-connected network, whose biases are zero(see Figure \ref{fig2}).
The $i$-th neuron at the $k$-th layer is represented as $x^k_i$ and the perturbed input is within $\mathbb{B}_\infty([0, 1]^T, 1)$.  
The input layer(orange) and the output layer(blue) both have two nodes, and the MaxPool function is a bivariate function for simplicity. 
In this example, $x^4_1$ is the output neuron of the true label.

Concretely, We get the value of $u^4_2$ by backsubstitution:
$$
\begin{aligned}
x^4_2&\leq -x_1^3+x_2^3\\
&\leq -x_1^2+0.5x_3^2+1\\
&\leq -x_1^1+0.5(0.25x_3^1+1.5)+1\\
% &\leq-0.93x_1^1+1.58\\
&\leq -x_1^0-x_2^0+0.5(0.25(2x_1^0-2x_2^0)+1.5)+1\\
&\leq -0.75x_1^0-1.25x_2^0+1.75\\
&\leq 2.5
\end{aligned}$$

Therefore, $u^4_2=2.5$. We get $l^4_1=-1, u^4_1=4.49, l^4_2=-2.99$ similarly.
$u^4_2 \geq l^4_1 $ means that $\epsilon=1$ is not a certified safe perturbation range, and we need to decrease $\epsilon$ to find the maximal robustness lower bound that we could certify.

\begin{center}
\begin{minipage}{8cm}
\begin{algorithm}[H]
\caption{Computing certified lower bound}
\label{alg:algorithm}
\begin{algorithmic}[1] %[1] enables line numbers
\REQUIRE  model F, input $\boldsymbol{x}$, true label $t$;\\
\ENSURE $\epsilon_{l}$;
\STATE Let $\epsilon_0\gets 0.005$,$\epsilon_{l}\gets \epsilon_0,\epsilon_{min}\gets 0,\epsilon_{max}\gets 1$.
\FOR{i=0 to 14} 
\STATE Compute $\boldsymbol{l^K}, \boldsymbol{u^K}$ of $F(\boldsymbol{x})$, where $\boldsymbol{x}\in\mathbb{B}_p(\boldsymbol{x_0},\epsilon_l)$
\IF {$l^K_t\geq max_{j\neq t}(u^K_j) $} 
\STATE $\epsilon_{min}=\epsilon_{l}$
\STATE $\epsilon_{l} =min(2\epsilon_{l},\frac{ \epsilon_{max}+\epsilon_{min}}{2}) $;
\ELSE 
\STATE $\epsilon_{max}=\epsilon_{l}$
\STATE $\epsilon_{l} =max(\frac{\epsilon_{l}}{2},\frac{ \epsilon_{max}+\epsilon_{min}}{2})$;
\ENDIF
\ENDFOR
\RETURN $\epsilon_{l}$
\end{algorithmic}
\end{algorithm}
\end{minipage}
\end{center}

\subsubsection{Computing maximal certified lower bound $\epsilon_{l}$}
 We use the binary search algorithm to find the maximal certified lower bound, which is the certified lower bound results in our work (see Algorithm \ref{alg:algorithm}). To make sure the perturbation range is larger than zero, we decrease or increase the perturbation range (lines 1, 6, and 9). When the perturbation range is certified safe (line 4), we then increase  $\epsilon$ (line 6); When $\epsilon$ cannot be certified safe, we then decrease  $\epsilon$ (line 9).  The difference between $\epsilon_{max}$ and $\epsilon_{min}$ is already reasonably small ($\leq2^{-15}$) after the process is repeated 15 times. Finally, after the above checking process is repeated 15 times,  the algorithm will terminate and return $\epsilon_l$ as the certified lower bound results. 
 For a K-layer convolutional network, if we assume that the $k$-th layer has $n_k$ neurons and the filter size is $k\times k$, the time complexity of MaxLin is $\mathcal{O}(K^2\times \max {n_k}^3)$. Detailed analysis are in the Appendix.

\section{Experimental Evaluation}\label{section4}

\begin{table*}
  \centering
  \scriptsize
  \caption{Averaged certified lower bounds and runtime on  CNNs on MNIST, CIFAR-10, and Tiny ImageNet datasets tested  by CNN-Cert, Ti-Lin, and MaxLin.}
  \resizebox{\textwidth}{!}{
  \begin{tabular}{c|c|c|rrr||rr||rrr}
  \hline
    \textbf{Dataset }  &  \textbf{Network  }    &   & \multicolumn{3}{c||}{\textbf{Certified Bounds($\boldsymbol{10^{-5}}$)}}  &\multicolumn{2}{c||}{\textbf{Bound Improvement(\%)}} &\multicolumn{3}{c}{\textbf{Average Runtime(min)} }\\\hline
          & & $\boldsymbol{l_p}$  & \multicolumn{1}{c|}{\textbf{CNN-Cert }}&\multicolumn{1}{c|}{\textbf{Ti-Lin }}& \multicolumn{1}{c||}{\textbf{MaxLin}}& \multicolumn{1}{r|}{\textbf{vs. CNN-Cert }}&\multicolumn{1}{r||}{\textbf{vs. Ti-Lin }}& \multicolumn{1}{c|}{\textbf{CNN-Cert}}& \multicolumn{1}{c|}{\textbf{  Ti-Lin  }  }&\textbf{  MaxLin  }     \\\hline
    
  \multirow{27}{*}{MNIST }& CNN&$l_\infty$&1318                &1837                &2083&\textbf{58.04$\uparrow$} &\textbf{13.39$\uparrow$}&  1.76&	1.73	&1.37 \\
    &   4 layers     & $l_2$&   4427                     & 6478           &7131&\textbf{61.08$\uparrow$} &   \textbf{10.08$\uparrow$}&1.39&	1.38&	1.40\\
  &36584 nodes & $l_1$  &  8544                   &12642   &13808&\textbf{61.61$\uparrow$} &\textbf{9.22$\uparrow$}&1.38&	1.38	&1.50\\\cline{2-11}

  &CNN & $l_\infty$ & 1288                &1817             &2712&\textbf{110.60$\uparrow$} &\textbf{49.26$\uparrow$}      &8.44	&8.76&	7.82\\
         & 5 layers     & $l_2$&  {5164} &7359   &9987&\textbf{93.40$\uparrow$} &\textbf{35.71$\uparrow$}           &11.90&	9.18&	7.47 \\
  &  52872 nodes & $l_1$ &    10147              &14292          &19000&\textbf{87.25$\uparrow$} &\textbf{32.94$\uparrow$}      &10.77&	9.46	&7.70 \\\cline{2-11}

    &CNN & $l_\infty$ &  1025&1382&1942&\textbf{89.46$\uparrow$} &\textbf{40.52$\uparrow$}  & 20.46	&20.87	&15.90  \\
       & 6 layers   & $l_2$&   3954                  &5409    &6981&\textbf{76.56$\uparrow$} &\textbf{29.06$\uparrow$}        &  20.56&	20.41&	15.94  \\
  &56392 nodes & $l_1$    & 7708              &10455 &13218&\textbf{71.48$\uparrow$} &\textbf{26.43$\uparrow$}      &20.60&	20.01&	15.93\\\cline{2-11}
   
   & CNN& $l_\infty$ &   647                  &930     &1289&\textbf{99.23$\uparrow$} &\textbf{38.60$\uparrow$}  &     24.71	&24.55&	18.91      \\
      &  7 layers     & $l_2$&    2733                     & 4022 &5228&\textbf{91.29$\uparrow$} &\textbf{29.99$\uparrow$} &  25.08&	23.80&	18.92            \\
  &56592 nodes& $l_1$ &    5443                     & 8002             &10248&\textbf{88.28$\uparrow$} &\textbf{28.07$\uparrow$}    &  22.86	&22.87	&18.78\\\cline{2-11}

  &CNN & $l_\infty$ &  847                     & 1221   &1666&\textbf{96.69$\uparrow$} &\textbf{36.45$\uparrow$}       &       26.51&	26.66	&22.19 \\
       & 8 layers     & $l_2$&     3751                 &5320        &6641&\textbf{77.05$\uparrow$} &\textbf{24.83$\uparrow$}      &  25.01&	24.85	&22.01    \\
   & 56912 nodes & $l_1$  &   7515                     & 10655         &12897&\textbf{71.62$\uparrow$} &\textbf{21.04$\uparrow$}        &    23.72	&24.23&	22.27     \\\cline{2-11}

  &LeNet\_ReLU   & $l_\infty$ &  1204&1864&2093& \textbf{73.83$\uparrow$}&\textbf{12.29$\uparrow$}  &0.16	&0.17&	0.17 \\
     & 3 layers    & $l_2$&  6534&10862&11750 & \textbf{79.83$\uparrow$}&\textbf{ 8.18$\uparrow$} &0.16&	0.17&	0.17 \\
  &8080 nodes& $l_1$  &  17937&30305&32313&\textbf{ 80.15$\uparrow$}&\textbf{ 6.63$\uparrow$}& 0.16	&0.17&	0.17\\\cline{2-11}

   &LeNet\_Sigmoid   & $l_\infty$ & 1684 &2042 &2567&\textbf{52.43$\uparrow$}&\textbf{25.71$\uparrow$}& 0.26&	0.28&	0.27\\
       & 3 layers   & $l_2$&9926&12369&14535&\textbf{46.43$\uparrow$}&\textbf{17.51$\uparrow$}&0.27&	0.27	&0.27\\
  &8080 nodes & $l_1$  &26937 &33384 &38264&\textbf{42.05$\uparrow$}&\textbf{14.62$\uparrow$}&0.27	&0.27&	0.27 \\\cline{2-11}

   & LeNet\_Tanh  & $l_\infty$ &613 &817&943&\textbf{53.83$\uparrow$}&\textbf{15.42$\uparrow$}&0.27	&0.27	&0.27  \\
    &3 layers    & $l_2$&3462&4916&5424&\textbf{56.67$\uparrow$}&\textbf{10.33$\uparrow$}&  0.27&	0.27	&0.27  \\
   &8080 nodes& $l_1$  &9566&13672 &14931&\textbf{56.08$\uparrow$}&\textbf{9.21$\uparrow$}&0.27	&0.27	&0.27 \\\cline{2-11}

   &  LeNet\_Atan   & $l_\infty$ &617&836 &961&\textbf{55.75$\uparrow$}&\textbf{14.95$\uparrow$}& 0.26&	0.27&	0.27 \\
   & 3 layers  & $l_2$&3514&5010 &5517&\textbf{57.00$\uparrow$}&\textbf{10.12$\uparrow$}&  0.28&	0.27	&0.27  \\
  &8080 nodes & $l_1$&9330&13345 &14522&\textbf{55.65$\uparrow$}&\textbf{8.82$\uparrow$}& 0.27	&0.28&	0.27\\\hline 
    \hline

  \multirow{15}{*}{CIFAR-10}&CNN & $l_\infty$&108                     &129      &147 &\textbf{36.11$\uparrow$} &\textbf{13.95$\uparrow$}& 3.09&	2.92	&2.94\\
         &4 layers    & $l_2$&     {751}&1038     &1172    &\textbf{56.06$\uparrow$}&\textbf{ 12.91$\uparrow$}  & 2.47&	2.51&	2.50\\
  &49320 nodes   & $l_1$  &    2127                     & 3029&3392 &\textbf{59.47$\uparrow$}&\textbf{11.98$\uparrow$}&2.46&	2.48	&2.49\\\cline{2-11}

   &CNN & $l_\infty$ &  115                     & 146 &169&\textbf{46.96$\uparrow$}&\textbf{15.75$\uparrow$}&13.10	&13.04	&13.07\\
   &     5 layers    & $l_2$&    953                     &1342&1519&\textbf{59.39$\uparrow$}&\textbf{13.19$\uparrow$}& 12.39&	12.69&	12.61\\
  &71880 nodes     & $l_1$ &     {2850}  & 4087               &4582 &\textbf{60.77$\uparrow$}&\textbf{12.11$\uparrow$}&12.34	&12.61&	12.51\\\cline{2-11}

  &CNN & $l_\infty$ &  99                     & 120     &139           &\textbf{40.40$\uparrow$} &\textbf{15.83$\uparrow$}     & 28.56&	28.63	&28.61  \\
        &6 layers    & $l_2$&     830                      &1078            &1217   &\textbf{46.63$\uparrow$}&\textbf{12.89$\uparrow$}&27.63	&27.89&	27.49 \\
   & 77576 nodes  & $l_1$    &   2387                     & 3174         &3558 &\textbf{49.06$\uparrow$}&\textbf{12.10$\uparrow$}    &27.66&	27.36&	27.68	 \\\cline{2-11}

   &CNN  & $l_\infty$ &  66                     & 83    &   96    &\textbf{45.45$\uparrow$}&  \textbf{15.66$\uparrow$} & 33.37	&33.27&	33.44 \\
        & 7 layers    & $l_2$&  573                  &773          &889&\textbf{55.15$\uparrow$}&\textbf{15.01$\uparrow$}    &  32.48&	32.77	&32.42  \\
   & 77776 nodes  & $l_1$ &     1673              &2303             &2623&\textbf{56.78$\uparrow$}&  \textbf{13.89$\uparrow$} &  33.56	&32.55	&32.96  \\\cline{2-11}

   &CNN & $l_\infty$ & 56                     &    70  &85&\textbf{51.79$\uparrow$}&\textbf{21.43$\uparrow$}    &   36.86&	37.54	&37.64    \\
       & 8 layers    & $l_2$&    536                     & 705   &835&\textbf{55.78$\uparrow$}&\textbf{18.44$\uparrow$}       & 37.46	&36.59	&36.91\\
  &  78416 nodes & $l_1$  &   1609                     & 2160     &2532&\textbf{57.36$\uparrow$}&\textbf{ 17.22$\uparrow$}        &36.89	&37.01&	37.38  \\\hline  \hline

    \multirow{3}{*}{  Tiny ImageNet}

  &CNN& $l_\infty$ &77 &123& 128   &\textbf{66.23$\uparrow$} &\textbf{ 4.07$\uparrow$} &184.94&183.98&185.81\\
       &      7 layers  & $l_2$&580&939& 962&\textbf{ 65.86$\uparrow$} &\textbf{ 2.45$\uparrow$} &184.36&183.25&185.07\\
 & 703512 nodes & $l_1$  &1747&2875 & 2934 &\textbf{67.95$\uparrow$} &\textbf{2.05$\uparrow$} &193.62 &183.93&184.03
\\  \hline  
       
  \end{tabular}}
  \label{table-cnncert}
  \end{table*}

In this section, we conduct extensive experiments on CNNs by comparing MaxLin with four state-of-the-art backsubstitution-based tools (CNN-Cert~\cite{boopathy2019cnn}, DeepPoly~\cite{singh2019abstract}, 3DCertify~\cite{lorenz2021robustness}, and Ti-Lin~\cite{xiao2022certifying}). 
Further, we compare MaxLin with BaB and multi-neuron abstraction tools ($\alpha$,$\beta$-CROWN~\cite{xu2020fast,wang2021beta,zhang2022general-gcp}, ERAN and MN-BaB~\cite{ferrari2022complete}).
The experiments run on a server running a 48 core Intel Xeon Silver 4310 CPU and 125 GB of RAM.

\subsection{Experimental Setup}

\textbf{Framework.}
The linear bounds of MaxLin are independent of the concrete verifier, and thus, we instantiate CNN-Cert~\cite{boopathy2019cnn} and 3DCertify~\cite{lorenz2021robustness} verifiers with MaxLin to certify the robustness of CNNs. Concretely, CNN-Cert verifier is the state-of-the-art verification framework and can support the $l_1, l_2, l_\infty$ perturbation form, while 3DCertify verifier is built atop ERAN framework and can certify various networks against  $l_\infty$ perturbation and other perturbation forms (such as rotation).

\textbf{Linear bounds for activations.}
As for the linear approximation of activations, 
we choose linear bounds in VeriNet~\cite{henriksen2020efficient}  as our Sigmoid/Tanh/Arctan's linear bounds. Further, we choose linear bounds in DeepPoly~\cite{singh2019abstract}  as our ReLU's linear bounds. These linear bounds are all the provable neuron-wise tightest~\cite{zhang2022provably} and stand for the highest precision among other relevant work~\cite{meng2022adversarial}. It is noticeable that when we compare MaxLin to other tools, only the linear bounds for MaxPool are different  for a fair comparison, that is, both the linear bounds of the activation functions and the other experiment setup are the same.

\textbf{Datasets.}
Our experiments are conducted on 
MNIST, CIFAR-10, and Tiny ImageNet, the well-known image datasets. The MNIST~\cite{lecun1998mnist} is a dataset of $28\times28$ handwritten digital images in 10 classes(from 0 to 9). 
CIFAR-10~\cite{krizhevsky2009learning} is a dataset of 60,000 $32\times32\times3$ images in 10 classes. 
Tiny ImageNet~\cite{deng2009imagenet} consists of 100,000  $64\times64\times3$  images in 200 classes. 
The value of each pixel is normalized into [0,1] and thus, the perturbation radius is in [0,1]. 

\textbf{Benchmarks.}
 We evaluate the performance of MaxLin on two classes of maxpool-based networks:  (I) CNNs, whose activation function is the ReLU function and with Batch Normalization; (II) LeNet, whose activation function is the Sigmoid, tanh, or arctan function. The networks used in experiments are all open-sourced and come from ERAN and CNN-Cert. 

\textbf{Metrics.} 
We refer to the metrics in CNN-Cert.
As for \textbf{tightness}, we use $\frac{100(\epsilon_{l}'-\epsilon_{l})}{\epsilon_{l}}\%$ to quantify the percentage of improvement, where $\epsilon_{l}'$ and $\epsilon_{l}$ represent the average certified lower bounds certified by MaxLin and other comparative tools, respectively. 
As for \textbf{efficiency}, we record the average computation time over the correctly-classified images and use $\frac{t}{t'}$ to represent the speedup of MaxLin over other baseline methods, where $t$ and $t'$ are the average computation time of MaxLin and other tools, respectively. Some detailed experiment setups are in  the Appendix.

\subsection{Performance on CNN-Cert}
As both MaxLin and Ti-Lin are built upon CNN-Cert, the state-of-the-art verification framework, we compare MaxLin to CNN-Cert and Ti-Lin.  
The generation way of the test set is the same as CNN-Cert, which generates 10 test images randomly. 

As for the tightness, MaxLin outperforms CNN-Cert and Ti-Lin in all settings with up to 110.60\% and 49.26\% improvement in Table \ref{table-cnncert}, respectively. The reason why MaxLin outperforms Ti-Lin, the neuron-wise tightest technique, is that minimizing the over-approximation zone is more effective for a single non-linear layer, whose nearest predecessor and posterior layers are linear. The MaxPool layer is usually placed after the activation layer and thus, Ti-Lin is inferior to MaxLin. 
As for time efficiency, as they share the same verification framework, CNN-Cert, and they can directly give linear bounds for MaxPool, the time cost of these three methods is almost the same.

\subsection{Performance on ERAN}

As MaxLin, DeepPoly, and 3DCertify are built upon the ERAN framework, which only can verify robustness against the $l_\infty$ adversary,
we compare MaxLin with DeepPoly and 3DCertify atop ERAN framework. 
CNNs with 4, 5, and 6 layers are from CNN-Cert, and CNNs with 7 and 8 layers are not supported by ERAN due to some undefined operations in the networks. MNIST\_LeNet\_Arctan is not used in this experiment as ERAN  does not support arctan. Furthermore, ERAN does not support Tiny ImageNet either.
The generation way of the test set is the same as ERAN, which chooses the first 10 images to test tools.

As for tightness, MaxLin outperforms DeepPoly and 3DCertify with up to 62.17\% and 39.94\% improvement in Table~\ref{table-3dcertify}, respectively.   MaxLin computes much tighter certified lower bounds than 3DCertify in most cases, and the only bad result of MaxLin only occurs in MNIST\_LeNet\_Sigmoid when compared with 3DCertify. This is reasonable. As the weights and biases of networks are quite different from each other, which makes the performance of verifiers varies on different networks as discussed in~\cite{zhang2022provably}.  However, we argue that MaxLin outperforms existing SOTA verifiers on MaxPool-based networks as MaxLin computes larger certified lower bounds in most cases.

\begin{table*}[t]
\centering
\caption{Averaged certified lower bounds and runtime on CNNs   on the MNIST and CIFAR-10 datasets tested  by  DeepPoly, 3DCertify,  and MaxLin.}

  \resizebox{\textwidth}{!}{
  \begin{tabular}{c|l|rrr||rr||rrr||r}
  \hline
    \textbf{Dataset  }  &  \multicolumn{1}{c|}{\textbf{Network  }}  &  \multicolumn{3}{c||}{\textbf{Certified Bounds($\mathbf{10^{-6}}$)}}& \multicolumn{2}{c||}{\textbf{Bound Improvement(\%)}}&\multicolumn{3}{c||}{\textbf{Average Runtime(min)} }&\multicolumn{1}{c}{\textbf{Speedup} }
    \\\hline
   & &  \multicolumn{1}{c|}{\textbf{DeepPoly}}&  \multicolumn{1}{c|}{\textbf{3DCertify }}& \multicolumn{1}{c||}{\textbf{MaxLin}} &  \multicolumn{1}{c|}{\textbf{vs. DeepPoly}}&  \multicolumn{1}{c||}{\textbf{vs. 3DCertify }}& \multicolumn{1}{c|}{\textbf{DeepPoly}} & \multicolumn{1}{c|}{\textbf{3DCertify}}  &\textbf{  MaxLin  }  &\multicolumn{1}{|c}{\textbf{vs.  3DCertify  } }    \\\hline
  
  \multirow{7}{*}{MNIST}&Conv\_Maxpool&2802&3247&4544 &\textbf{62.17$\uparrow$}&\textbf{39.94$\uparrow$}&0.54	&1.21	&0.58&\textbf{2.09}\\\cline{2-11}

  &CNN, 4 layers&9375&10621&11272& \textbf{20.23$\uparrow$}&\textbf{6.13$\uparrow$}&1.34	&4.44&	1.48&\textbf{3.01}\\\cline{2-11}
  
  &CNN, 5 layers&6642&7629&7948&\textbf{19.66$\uparrow$} &\textbf{4.18$\uparrow$}&5.40&	13.13&	5.51&\textbf{2.38}\\\cline{2-11}
  
  &CNN, 6 layers&6339&7325&7554& \textbf{19.17$\uparrow$}&\textbf{3.13$\uparrow$}&11.88	&27.87&	12.47&\textbf{2.23}\\\cline{2-11}
  
  &LeNet\_ReLU&8849&10937&11225& \textbf{26.85$\uparrow$}&\textbf{2.63$\uparrow$}&0.14	&0.69&	0.19&\textbf{3.70}\\\cline{2-11}
  
  &LeNet\_Sigmoid&12122&14716&14506&\textbf{19.67$\uparrow$}&-1.43$\downarrow$&0.15	&1.01	&0.20&\textbf{5.13}\\\cline{2-11}

  &LeNet\_Tanh&2966&3637&3675& \textbf{23.90$\uparrow$}&\textbf{1.04$\uparrow$}& 0.17&	0.82&0.19&\textbf{4.31}\\\cline{2-11}
  
  \hline\hline
  
  \multirow{4}{*}{CIFAR-10}&Conv\_Maxpool&661&725& 754&\textbf{14.07$\uparrow$}&\textbf{4.00$\uparrow$}&8.16	&9.84&	8.35&\textbf{1.18}\\\cline{2-11}
  
  &CNN, 4 layers&1204&1460&1542&\textbf{ 28.07$\uparrow$}&\textbf{5.62$\uparrow$}&2.64	&4.25&	2.64&\textbf{1.61}\\\cline{2-11}
  
  &CNN, 5 layers&1223&1537&1579&\textbf{ 29.11$\uparrow$}&\textbf{2.73$\uparrow$}&11.82	&17.75	&12.44&\textbf{1.43}\\\cline{2-11}
  
  &CNN, 6 layers&1065&1415&1440&\textbf{ 35.21$\uparrow$}&\textbf{1.77$\uparrow$}&24.60&	40.44&	24.89&\textbf{1.62}\\\hline
  
  \end{tabular}}
  \label{table-3dcertify}
  \end{table*}

As for time efficiency, 3DCertify is quite time-consuming as it tries to find the best upper linear bound from the linear bounds set gained 
by the Double Description Method~\cite{fukuda2005double}.
However, MaxLin can give the upper and lower linear bounds directly after choosing the first and second maximum values of the upper and lower bound of maxpool's input  $\boldsymbol{l}$ and $\boldsymbol{u}$ and thus, is efficient. Therefore, MaxLin has up to 5.13$\times$ speedup compared with 3DCertify and  almost the same time efficiency as DeepPoly in Table~\ref{table-3dcertify}. 

\subsection{Evaluating The Block-wise Tightness}
To further illustrate the superiority of the block-wise tightness,
we compare MaxLin and the baselines by the volume of the Activation+MaxPool block. 
The pool size is $2\times2$, and the number of inputs is 50.  The Activation has three types: 
(i) ReLU, whose linear bounds are the provably neuron-wise tightest~\cite{singh2019abstract,zhang2018efficient}; 
(ii) Adaptive-ReLU~\cite{xu2020fast}, whose upper linear bounds is $u(x)=\frac{ReLU(u)-ReLU(l)}{u-l}$ and lower linear bounds is adaptive: $l(x)=ax, a\in[0,1]$; 
(iii) Sigmoid, whose linear bounds are the provably neuron-wise tightest~\cite{henriksen2020efficient}.  
Specifically, we employ a random sampling approach to determine both the upper and lower bounds for each pixel, following a uniform distribution $U(-10, 10)$. Simultaneously, we randomly select the value of $a$ from a uniform distribution $U(0, 1)$.
\begin{figure} 
\centering
\includegraphics[width=1.0\columnwidth]{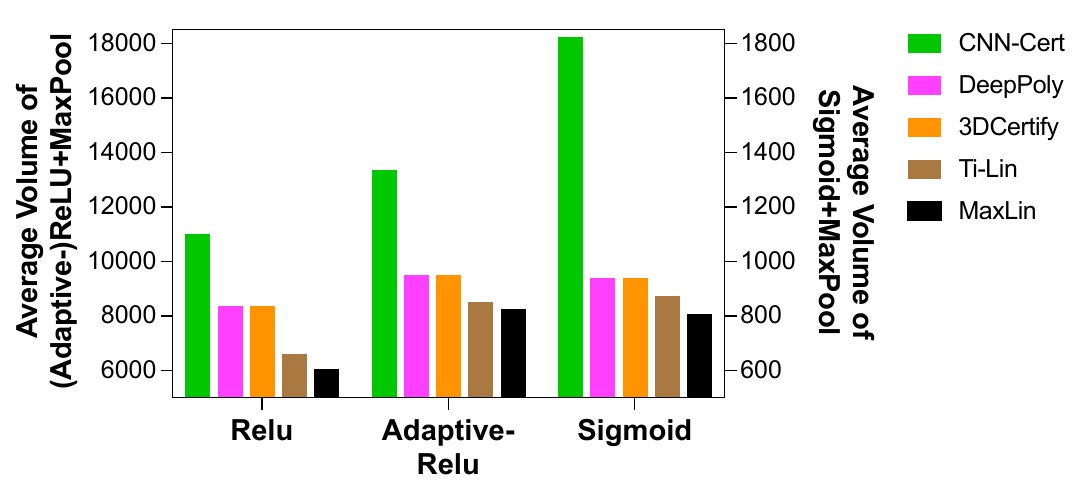} 
\caption{The average volume 
of the Activation+MaxPool block computed by CNN-Cert, DeepPoly, 3DCertify, Ti-Lin and MaxLin.}
\label{fig3}
\end{figure}
Figure~\ref{fig3} shows the average volume 
of the Activation+MaxPool block computed by the baselines and MaxLin.
Concretely, MaxLin has the smallest volume of the over-approximation zone of the ReLU+MaxPool  and Adaptive-ReLU+MaxPool blocks among the baseline methods, which validates the correctness of Theorem~\ref{theorem provable tightest}. 
Further,  in terms of S-shaped activation functions, MaxLin has the smallest results regarding the average volume. 
This shows that when the upper linear bound is not the provably block-wise tightest, MaxLin can also reduce the over-approximation zone of the non-linear block. Moreover, The results show that the neuron-wise tightest linear bounds (Ti-Lin) could only keep high precsion through one layer, while MaxLin could keep the tightness through one-block propagation. 
The results in Figure~\ref{fig3} are consistent with the results in Table~\ref{table-cnncert} and~\ref{table-3dcertify} and indicate the advantage of the block-wise tightest upper linear bound in terms of precision.

\subsection{Additional Experiments}
We conduct additional experiments to further demonstrate the superiority and broad applicability of MaxLin.
The detailed settings are in the Appendix, and we perform the following experiments:
(I) We compare the output interval $[\boldsymbol{l^K}, \boldsymbol{u^K}]$ computed by MaxLin and Ti-Lin to further illstrates the advantages of the block-wise tightness over the neuron-wise tightness.
(II) We conduct extensive experiments by comparing the time efficiency of BaB-based and backsubstitution-based verification frameworks. 
(III) We compare MaxLin with BaB-based verification frameworks, including VNN-COMP 2021-2023~\cite{bak2021second,muller2022third} winner $\alpha$,$\beta$-CROWN~\cite{xu2020fast,wang2021beta,zhang2022general-gcp}, ERAN using multi-neuron abstraction and MN-BaB~\cite{ferrari2022complete} on ERAN benchmark. 
(IV) We also conduct experiments by certifying the robustness of PointNet on the ModelNet40 dataset~\cite{wu20153d}  to illustrate the broad applicability of MaxLin.

\section{Conclusion}
In this paper, we propose MaxLin, a tight linear approximation approach to MaxPool for computing larger certified lower bounds for CNNs. 
MaxLin has high execution efficiency as it uses the single-neuron relaxation technique and computes linear bounds with low computational consumption. MaxLin is built atop CNN-Cert and 3DCertify, two state-of-the-art verification frameworks, and thus, can certify the robustness of various networks(e.g., CNNs and PointNet) with arbitrary activation functions against $l_1, l_2, l_\infty$ perturbation form. 
We evaluate MaxLin with open-sourced benchmarks on the MNIST, CIFAR-10, and Tiny ImageNet datasets. The results show that MaxLin outperforms the SOTA tools with at most 110.60\% improvement regarding the certified lower bound and 5.13 $\times$ speedup for the same neural networks.

{
    \small
    \bibliographystyle{ieeenat_fullname}
    \bibliography{reference}
}

\newpage
\section{Appendix}

\subsection{Experiment Setups}\label{sec:expsetting}

In this subsection, we present some experiment setups in detail. Concretely, we list the sources of networks, the size of the test set, and the initial perturbation range in Table \ref{table-network}.  We evaluate methods on 10 inputs for all CNNs in Table~\ref{table-cnncert} and \ref{table-3dcertify}. As CIFAR\_Conv\_MaxPool has low accuracy and low robustness, the size of the test set is 50 and the initial perturbation range is 0.00005. 
Testing on 10 inputs can sufficiently evaluate the performance of verification methods, as it is shown that the average certified results of 1000 inputs are similar to 10 images~\cite{boopathy2019cnn}. 
\subsection{Additional Experiments}

In this subsection, we conduct some additional experiments to further illustrate (I) the advantage of the block-wise tighness over the neuron-wise tightness, (II) the time efficiency of MaxLin compared to other BaB-based verification tools, (III) the performance of MaxLin using multi-neuron abstraction techniques, and (IV) the performance of MaxLin on PointNets. 

We list the sources of networks for additional experiments in Table~\ref{table-network}. We evaluate methods on 100 inputs for ERAN benchmark (CIFAR\_Conv\_MaxPool) and 100 inputs for PointNets. The perturbation range $\epsilon$ is 0.005 for PointNets and is 0.0007, 0.0008, 0.0009, 0.0010, or 0.0011 for CIFAR\_Conv\_MaxPool.   
We follow the metrics used in the baseline methods.
As for the effectiveness, we use certified accuracy, the percentage of the successfully verified inputs against the perturbation range, to evaluate the tightness of methods. We also use the average per-example verified time as the metric for time efficiency. In additional experiments II and III, we compare MaxLin with three state-of-the-art verification techniques, which use the BaB and multi-neuron abstraction technique to enhance the precision. Concretely, the baselines are MN-BaB~\cite{ferrari2022complete}, $\alpha$,$\beta$-CROWN~\cite{xu2020fast,wang2021beta,zhang2022general-gcp} (VNN-COMP 2021~\cite{bak2021second} and 2022~\cite{muller2022third} winner), and ERAN.  In additional experiment IV, we evaluate the performance of MaxLin and three single-neuron abstraction methods(DeepPoly~\cite{singh2019abstract}, 3DCertify~\cite{lorenz2021robustness}, and Ti-Lin~\cite{xiao2022certifying}) on PointNets.

\begin{table}[htbp]
\centering
\renewcommand\arraystretch{1.3}
\caption{The additional experimental setup and source  of neural networks used in experiments. $\epsilon_0$ is the initial perturbation range in Algorithm \ref{alg:algorithm}. The third column represents the size of the input test set. }
\resizebox{0.98\columnwidth}{!}{
\begin{tabular}{c|l|c|r|c}
\hline \textbf{Dataset  }  &  \textbf{Network}  & \multicolumn{1}{c|}{ \textbf{Size}} &\multicolumn{1}{c}{{$\boldsymbol{\epsilon_0}$}}&\multicolumn{1}{|c}{\textbf{Source}} \\\hline

\multirow{10}{*}{MNIST}&Conv\_MaxPool&10&0.005&ERAN \\\cline{2-5}

&CNN, 4 layers & 10&0.005& \multirow{8}{*}{CNN-Cert}\\\cline{2-4}

&CNN, 5 layers &10&0.005& \\\cline{2-4}

&CNN, 6 layers &10&0.005& \\\cline{2-4}

&CNN, 7 layers &10&0.005& \\\cline{2-4}

&CNN, 8 layers &10&0.005& \\\cline{2-4}

&LeNet\_ReLU&10&0.005&\\\cline{2-4}

&LeNet\_Sigmoid&10&0.005&\\\cline{2-4}

&LeNet\_Tanh&10&0.005&\\\cline{2-4}
&LeNet\_Atan&10&0.005&\\\cline{2-5}

\hline\hline
\multirow{6}{*}{CIFAR-10}&Conv\_MaxPool&50 & 0.00005&ERAN\\\cline{2-5}

&CNN, 4 layers &10&0.005&\multirow{5}{*}{CNN-Cert}\\\cline{2-4}

&CNN, 5 layers &10&0.005& \\\cline{2-4}
&CNN, 6 layers &10&0.005&\\\cline{2-4}
&CNN, 7 layers &10&0.005&\\\cline{2-4}

&CNN, 8 layers &10&0.005&\\

\hline
\hline
\multirow{1}{*}{Tiny ImageNet}

&CNN, 7 layers&10&0.005&\\\hline\hline
\multirow{5}{*}{ModelNet40}&16p\_natural&100&0.005&\multirow{5}{*}{3DCertify}\\\cline{2-4}

&32p\_natural&100&0.005&\\\cline{2-4}
&64p\_natural&100&0.005&\\\cline{2-4}
&128p\_natural&100&0.005&\\\cline{2-4}
&256p\_natural&100&0.005&\\

\hline

\end{tabular}}
\label{table-network}
\end{table}

\begin{figure}[!t]
\centering

\includegraphics[width=1.0\columnwidth]{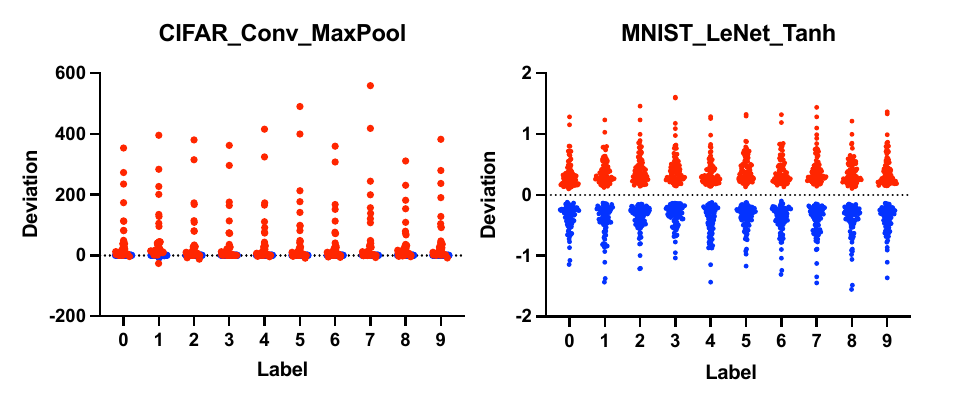} 
\caption{Visualization of the output intervals verified by MaxLin and Ti-Lin.
$(\boldsymbol{l},\boldsymbol{u})$ and $(\boldsymbol{l'},\boldsymbol{u'})$ represent the output bound of Ti-Lin and MaxLin testing on 100 inputs, respectively. Red and blue dots represent $\boldsymbol{u}-\boldsymbol{u'}$ and  $\boldsymbol{l}-\boldsymbol{l'}$, respectively.}
\label{fig7}
\end{figure}

\subsubsection{Results (I): Advantages Over The Neuron-wise Tightest Method}
To further illustrate the advantage of the block-wise tightness (MaxLin) over the neuron-wise tightness(Ti-Lin), we analyze the verified interval  of the output neurons of the last layer. In Figure~\ref{fig7}, we compare the results computed by MaxLin and Ti-Lin on CIFAR\_Conv\_MaxPool and MNIST\_LeNet\_Tanh, whose actvations are ReLU and S-shaped, respectively. For the output neurons(10 labels), we use $(\boldsymbol{l},\boldsymbol{u})$ and $(\boldsymbol{l'},\boldsymbol{u'})$ to represent the output bound of Ti-Lin and MaxLin testing on 100 inputs, respectively. We use the red and blue dots to represent $\boldsymbol{u}-\boldsymbol{u'}$ and  $\boldsymbol{l}-\boldsymbol{l'}$, respectively. The $x$-axis represents the output neuron index(label), and the $y$-axis represents the deviations between the  lower and upper bounds of the intervals. 

As the activation of CIFAR\_Conv\_MaxPool is ReLU, the lower bounds of the ouput neurons are mostly zero and thus the deviation of lower bounds is zero. Except for this case, most $u-u'$ are larger than zero and $l-l'$ are smaller than zero in Figure~\ref{fig7}. It reveals that the block-wise tightest upper linear bounds could bring tighter output intervals than the neuron-wise tightest linear bounds. Consequently, MaxLin could certify much larger robustness bounds than Ti-Lin in Table~\ref{table-cnncert}.

\subsubsection{Results (II): Performance Using Single-neuron Abstraction for ReLU}\label{sec:exp1} 

\begin{figure} 
\centering
\includegraphics[width=1.0\columnwidth]{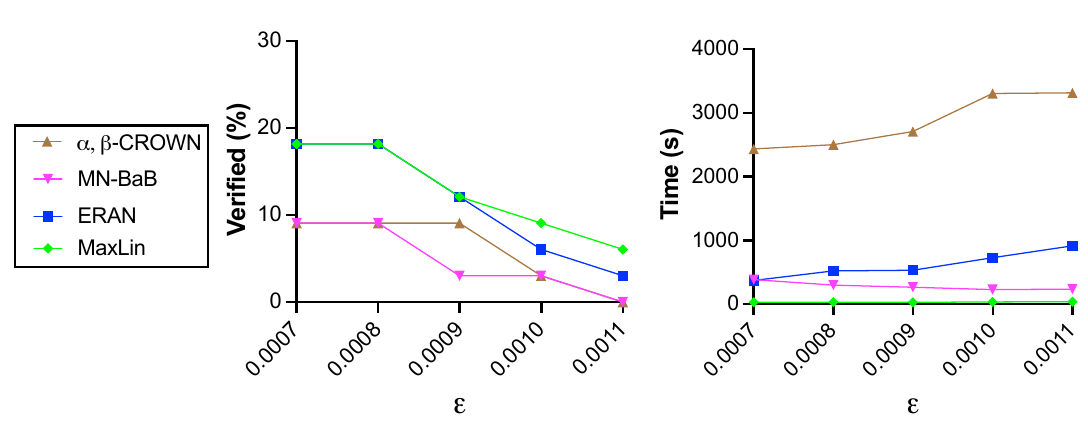} 
\caption{Certified accuracy(\%) and average per-example verification time(s) on CIFAR\_Conv\_MaxPool tested by MN-BaB, $\alpha$,$\beta$-CROWN, ERAN, and MaxLin(using single-neuron techniques for ReLU).}
\label{fig3d}
\end{figure}

We conduct additional experiments to present the time efficiency of the single-neuron abstraction technique, which is used by MaxLin in Section~\ref{section4}. We compare MaxLin(single-neuron abstraction for ReLU) with three state-of-the-art verification techniques(MN-BaB~\cite{ferrari2022complete}, $\alpha$,$\beta$-CROWN~\cite{xu2020fast,wang2021beta}, and ERAN using multi-neuron abstraction) on CIFAR\_Conv\_MaxPool. The results are presented in Figure~\ref{fig3d}. In terms of time efficiency, MaxLin can accelerate the computation process with up to $14.1,  96.5,$ and $23.8\times$ compared to MN-BaB, $\alpha,\beta$-CROWN, and ERAN, respectively. Although MaxLin uses the single-neuron abstraction technique for ReLU, MaxLin still can enhance precision with up to 9.1, 9.1, and 3.0\% improvement compared to MN-BaB, $\alpha,\beta$-CROWN, and ERAN , respectively. These results demonstrate that the single-neuron abstraction technique has the potential of verifying large models and other complex models, such as PointNets(results in Subsection~\ref{sec:exp3})

\subsubsection{Results (III): Performance Using Multi-neuron Abstraction for ReLU}\label{sec:exp2} 
\begin{figure} 
\centering
\includegraphics[width=1.0\columnwidth]{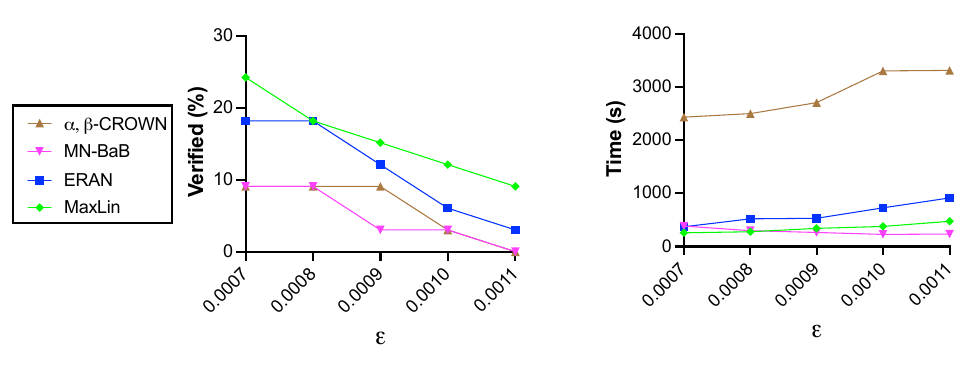} 
\caption{Certified accuracy(\%) and average per-example verification time(s) on CIFAR\_Conv\_MaxPool tested by MN-BaB, $\alpha$,$\beta$-CROWN, ERAN, and MaxLin(using multi-neuron techniques for ReLU).}
\label{figeran}
\end{figure}

As ERAN framework, atop which MaxLin is built,   not only supports single-neuron abstraction but also integrates the multi-neuron abstraction for ReLU. We compare MaxLin using multi-neuron abstraction for ReLU to MN-BaB, $\alpha$, $\beta$-CROWN, and ERAN using multi-neuron abstraction on CIFAR\_Conv\_MaxPool. The results are shown in Figure~\ref{figeran}. The results show that MaxLin has higher certified accuracy with up to 15.2, 15.2, 6.1\% improvement compared to MN-BaB, $\alpha$, $\beta$-CROWN, and ERAN, respectively. In terms of time efficiency, MaxLin has similar time cost to MN-BaB and has up to 9.5 and 1.9 $\times$ speedup compared to $\alpha$, $\beta$-CROWN and ERAN, respectively.
These results show that if using multi-neuron abstraction, MaxLin also could have higher  certified accuracy and less time cost than these verification tools.

\begin{table*}[htbp]
\centering
\caption{Averaged certified results and runtime on PointNet  on the ModelNet40 datasets tested  by  DeepPoly, 3DCertify, Ti-Lin, and MaxLin, where 16p\_natural represents the PointNet model is naturally trained and the number of its point input  is 16. }
\renewcommand\arraystretch{1.5}
  \resizebox{\textwidth}{!}{
  \begin{tabular}{c|l||rrrr||rrrr||r}
  \hline
   \multicolumn{1}{c|}{\textbf{Dataset}}&  \multicolumn{1}{c||}{\textbf{Network }}  &  \multicolumn{4}{c||}{\textbf{Certified accuracy (\%)}}&\multicolumn{4}{c||}{\textbf{Average Runtime(second)} }&\multicolumn{1}{c}{\textbf{Speedup} }
    \\\hline
  & &    \multicolumn{1}{c|}{\textbf{ DeepPoly }} &  \multicolumn{1}{c|}{\textbf{ 3DCertify }} & \multicolumn{1}{c|}{\textbf{Ti-Lin}}& \multicolumn{1}{c||}{\textbf{MaxLin}}&\multicolumn{1}{c|}{\textbf{DeepPoly}}  & \multicolumn{1}{c|}{\textbf{3DCertify}}& \multicolumn{1}{c|}{\textbf{Ti-Lin}}  &\textbf{  MaxLin  }  &\multicolumn{1}{c}{\textbf{vs.  3DCertify  } }    \\\hline

\multirow{5}{*}{ModelNet40}
&16p\_natural&72.73&\textbf{74.03}&\textbf{74.03}&\textbf{74.03} &10.37&18.22&12.47&10.61&\textbf{1.72}\\\cline{2-11}
&32p\_natural&54.88&58.54&58.54&\textbf{64.63}&17.88&34.79&21.24&18.06&\textbf{1.93}\\\cline{2-11}
&64p\_natural&32.56&40.70&36.05&\textbf{47.67}&35.85&96.04&42.96&36.63&\textbf{2.62}\\\cline{2-11}
&128p\_natural&4.55&11.36&2.27&\textbf{14.77}&81.46&207.04&110.16&85.90&\textbf{2.41}\\\cline{2-11}
&256p\_natural&1.12&\textbf{4.49}&1.12&\textbf{4.49}&178.34&494.70&248.19&199.88&\textbf{2.47}\\\cline{2-11}
\hline
\end{tabular}}
\label{table-pointnet-perturbation}
\end{table*}

\subsubsection{Results (IV): Performance on PointNets}\label{sec:exp3}

3D point cloud models are widely used and achieve great success in some safety-critical domains, such as autonomous driving. It is of vital importance to provide a provable robustness guarantee to models before deployed.

As 3DCertify  is a robustness verifier for point cloud models,
MaxLin,  which is  built atop the  3DCertify framework, can be  extended to  certify the robustness of 3D Point Cloud models against point-wise $l_\infty$ perturbation and 3D transformation. Here, we demonstrate that MaxLin is not only useful beyond image classification models but also performs well on other models. To that end, we show certification results against point-wise $l_\infty$ perturbation on seven PointNets for the ModelNet40~\cite{wu20153d} dataset in Table \ref{table-pointnet-perturbation}. The PointNet whose inputs' point number is $k$ is denoted as $k$p\_natural. 
As Ti-Lin is the best state-of-the-art tool built on the CNN-Cert framework, we integrate its linear bounds for MaxPool  into the 3DCertify framework as one baseline. The generation way of the test set is the same as 3DCertify and all experiments use the same random subset of 100 objects from the ModelNet40~\cite{wu20153d} dataset. The certification results represent the percentage of verified robustness properties and the perturbation range is 0.005. The perturbation is in $l_\infty$ norm and is measured by Hausdroff distance. 

As for tightness, the certification results in Table \ref{table-pointnet-perturbation} show that 
MaxLin outperforms other tools in all cases in terms of tightness. It is reasonable that the results for 16p\_natural and 256\_natural  certified by  3DCertify and MaxLin are the same, as the perturbation range is not large enough to distinguish the tightness of these tools.
As for efficiency, in Table \ref{table-pointnet-perturbation}, MaxLin has up to 2.62 $\times$ speedup compared with 3DCertify and is slightly faster than Ti-Lin. MaxLin has almost the same time consumption as DeepPoly.
In summary,  these results demonstrate that our fine-grained linear approximation can help improve both the tightness and efficiency of robustness verification of other models beyond the image classification domain.

\subsection{Complexity analysis of MaxLin}
For a K-layer convolutional network, we assume that the $k$-th layer has $n_k$ neurons and the filter size is $k\times k$.
   The time complexity of backsubstitution is $\mathcal{O}(K\times\max n_k^3)$~\cite{zhang2018efficient} and the backsubstitution process will be repeated $ K-1$ times to verify one input perturbed within a certain perturbation range. Therefore,  the time complexity of MaxLin is $\mathcal{O}(K^2\times \max {n_k}^3)$.

\subsection{Proof of Theorem \ref{theoremMaxPool}}

We prove the correctness of linear bounds in Theorem \ref{theoremMaxPool} as follows.

\begin{proof}

\textbf{Upper linear bound:}

Case 1:

When $(l_i=l_{max})\wedge(l_i\geq u_j)$, $u(x_1,\cdots,x_n)=x_i$ and $f(x_1,\cdots,x_n)=x_i$.  Therefore,

$$\begin{aligned}u(x_1,\cdots,x_n)-f(x_1,\cdots,x_n)=&x_i-x_i\\
=&0\end{aligned}$$

Case 2:

Otherwise, $f(x_1,\cdots,x_n)=max\{x_1,\cdots,x_n\}$

If $f(x_1,\cdots,x_n)=x_i$,
$$\begin{aligned}&u(x_1,\cdots,x_n)-f(x_1,\cdots,x_n)\\
=&\frac{u_i-u_j}{u_i-l_i}(x_i-l_i)+u_j-x_i\\
=&\frac{l_i-u_j}{u_i-l_i}x_i-\frac{u_j-l_i}{u_i-l_i}u_i\\
=&\frac{u_j-l_i}{u_i-l_i}(u_i-x_i)\\
\geq&0
\end{aligned}$$

If $f(x_1,\cdots,x_n)=x_j$,
$$\begin{aligned}&u(x_1,\cdots,x_n)-f(x_1,\cdots,x_n)\\
=&\frac{u_i-u_j}{u_i-l_i}(x_i-l_i)+u_j-x_j\\
\geq&0
\end{aligned}$$

Otherwise, we assume $f(x_1,\cdots,x_n)=x_q$, where $q\neq i,j$ and $q\in[n]$.
$$\begin{aligned}&u(x_1,\cdots,x_n)-f(x_1,\cdots,x_n)\\
=&\frac{u_i-u_j}{u_i-l_i}(x_i-l_i)+u_j-x_q\\
\geq&\frac{u_i-u_j}{u_i-l_i}(x_i-l_i)+u_j-u_q\\
\geq&0
\end{aligned}$$

\textbf{Lower linear bound:}
$$\begin{aligned}f(x_1,\cdots,x_n)=&max(x_1,\cdots,x_n)\\\geq& x_j\\=&l(x_1,\cdots,x_n)\\\end{aligned}$$

This completes the proof.  
\end{proof}

\subsection{Proof of Theorem \ref{theorem provable tightest}}
First,  we prove minimizing the volume of the over-approximation zone of the linear bounds $U^{k+1}_b(\cdot)$ and $L^{k+1}_b(\cdot)$ for non-linear block is equivalent to minimizing $U^{k+1}_b(m^{k-1})$ and $L^{k+1}_b(m^{k-1})$, respectively.

\begin{theorem}
\label{theorem equival}

     Minimizing 
     $$\iint_{\boldsymbol{x^{k-1}}\in[\boldsymbol{l^{k-1},\boldsymbol{u^{k-1}]}}}(U^{k+1}_b(\boldsymbol{x^{k-1}})-L^{k+1}_b(\boldsymbol{x^{k-1}}))d\boldsymbol{x^{k-1}}$$   
     is equivalent to minimizing  $U^{k+1}_b(\boldsymbol{m^{k-1}})$ and $L^{k+1}_b(\boldsymbol{m^{k-1}})$
\end{theorem}

The proof of Theorem \ref{theorem equival} is as follows.
\begin{proof}
First, Minimizing 
$$\iint_{\boldsymbol{x^{k-1}}\in[\boldsymbol{l^{k-1},\boldsymbol{u^{k-1}]}}}(U^{k+1}_b(\boldsymbol{x^{k-1}})-L^{k+1}_b(\boldsymbol{x^{k-1}}))d\boldsymbol{x^{k-1}}$$ 
is equivalent to minimizing the 
$$\iint_{\boldsymbol{x^{k-1}}\in[\boldsymbol{l^{k-1},\boldsymbol{u^{k-1}]}}}(U^{k+1}_b(\boldsymbol{x^{k-1}})-f^{k-1}(\boldsymbol{x^{k-1}}))d\boldsymbol{x^{k-1}}$$ 
    and 
    $$\iint_{\boldsymbol{x^{k-1}}\in[\boldsymbol{l^{k-1},\boldsymbol{u^{k-1}]}}}(f^{k-1}(\boldsymbol{x^{k-1}})-L^{k+1}_b(\boldsymbol{x^{k-1}}))d\boldsymbol{x^{k-1}}$$, respectively.

Therefore, it is equivalent to minimize the $\iint_{\boldsymbol{x^{k-1}}\in[\boldsymbol{l^{k-1},\boldsymbol{u^{k-1}]}}}U^{k+1}_b(\boldsymbol{x^{k-1}})d\boldsymbol{x^{k-1}}$ 
    and 
    $\iint_{\boldsymbol{x^{k-1}}\in[\boldsymbol{l^{k-1},\boldsymbol{u^{k-1}]}}}(-L^{k+1}_b(\boldsymbol{x^{k-1}}))d\boldsymbol{x^{k-1}}$, respectively.

Because $U^{k+1}_b(\boldsymbol{x^{k-1}})$ is a linear combination of $\boldsymbol{x^{k-1}}$. Without loss of generality, we assume $U^{k+1}_b(\boldsymbol{x^{k-1}})=\sum_{q\in[n_{k-1}]}A_{u,q}^{k+1} x_q^{k-1}+B^{k+1}_{u}$. Then,
$$
\begin{aligned}
&\iint_{\boldsymbol{x}\in[\boldsymbol{l^{k-1}},\boldsymbol{u^{k-1}}]}U^{k+1}_b(\boldsymbol{x^{k-1}})d\boldsymbol{x}\\
=&\iint_{\boldsymbol{x}\in[\boldsymbol{l^{k-1}},\boldsymbol{u^{k-1}}]}(\sum_{q\in[n_{k-1}]}A_{u,q}^{k+1} x_q+B^{k+1}_{u})d\boldsymbol{x}\\
=&\Pi_{i=1}^{n_{k-1}}(u^{k-1}_i-l^{k-1}_i)(\sum_{q\in[n_{k-1}]}A_{u,q}^{k+1}\frac{u^{k-1}_q+l^{k-1}_q}{2}+B^{k+1}_{u})\\
=&\Pi_{i=1}^{n_{k-1}}(u^{k-1}_i-l^{k-1}_i)U_b^{k+1}(\boldsymbol{m^{k-1}})\\
\end{aligned}\nonumber
$$

where $\boldsymbol{m^{k-1}}=(\frac{u^{k-1}_1+l^{k-1}_1}{2},\cdots,\frac{u^{k-1}_{n_{k-1}}+l^{k-1}_{n_{k-1}}}{2})$, because $u^{k-1}_q,l^{k-1}_q$, $q\in[n_{k-1}]$ are constant, and the minimize target has been transformed into minimizing $U_b^{k+1}(\boldsymbol{m})$.

Therefore, minimizing $\iint_{\boldsymbol{x^{k-1}}\in[\boldsymbol{l^{k-1},\boldsymbol{u^{k-1}]}}}U^{k+1}_b(\boldsymbol{x^{k-1}})$ 
    is equivalent to minimize  $U^{k+1}_b(\boldsymbol{m^{k-1}})$ 

   Similarly, minimizing 
    $-\iint_{\boldsymbol{x^{k-1}}\in[\boldsymbol{l^{k-1},\boldsymbol{u^{k-1}]}}}L^{k+1}_b(\boldsymbol{x^{k-1}})$  is equivalent to minimize  $-L^{k+1}_b(\boldsymbol{m^{k-1}})$
\end{proof}

Based on Theorem~\ref{theorem equival}, we prove Theorem \ref{theorem provable tightest} as follows.
\begin{proof}
When $l<0\wedge u>0$, the linear bounds of ReLU used in our approach are the same as~\cite{singh2019abstract,zhang2018efficient,boopathy2019cnn}, which are the provable neuron-wise tightest upper linear bounds~\cite{singh2019abstract}. It is:
    \begin{equation}
    \begin{aligned}
                &u(x)=\frac{u}{u-l}(x-l)
    \end{aligned}
    \end{equation}

First, we prove the upper linear bound is the block-wise tightest. Without loss of generality, we assume ReLU is at the $k$-th layer.  We use $u^{k+1},l^{k+1}$ and $u_M^{k+1},l_M^{k+1}$ to denote other linear bounds and our linear bounds for the MaxPool function, respectively.
As the slope of the MaxPool linear bounds is always non-negative, the global upper and lower linear bounds of the ReLU+MaxPool block are:
$$\begin{aligned}
    &u^{k+1}(x^k_1,\cdots,x^k_n)\\
    \leq& u^{k+1}(u^{k}(x^{k-1}_1),\cdots,u^{k}(x^{k-1}_n))\\
    =:&U^{k+1}_b(x^{k-1}_1,\cdots,x^{k-1}_n)\\
    &l^{k+1}(x^k_1,\cdots,x^k_n)\\
    \leq& l^{k+1}(l^{k}(x^{k-1}_1),\cdots,l^{k}(x^{k-1}_n))\\
    =:&L^{k+1}_b(x^{k-1}_1,\cdots,x^{k-1}_n)\\
\end{aligned}
$$
where we use $U^{k+1}_b(\cdot)$ and $L_b^{k+1}(\cdot)$ to denote the global upper and lower linear bounds of the ReLU+MaxPool block, respectively.

If $U^{k+1}_b(m^{k-1}_1,\cdots,m^{k-1}_n)$ reaches its minimum, the upper linear bound of Theorem \ref{theoremMaxPool} are the provable block-wise tightest.

\textbf{Upper linear bound:}

$$
\begin{aligned}
    &U^{k+1}_b(m^{k-1}_1,\cdots,m^{k-1}_n)\\
    =&u^{k+1}(u^{k}(m^{k-1}_1),\cdots,u^{k}(m^{k-1}_n))\\
=&u^{k+1}(\frac{u^{k-1}_1(m^{k-1}_1-l^{k-1}_1)}{u^{k-1}_1-l^{k-1}_1},\cdots,\frac{u^{k-1}_n(m^{k-1}_n-l^{k-1}_n)}{u^{k-1}_n-l^{k-1}_n})\\
=&u^{k+1}(\frac{u^{k-1}_1(\frac{u^{k-1}_1-l^{k-1}_1}{2})}{u^{k-1}_1-l^{k-1}_1},\cdots,\frac{u^{k-1}_n(\frac{u^{k-1}_1-l^{k-1}_1}{2})}{u^{k-1}_n-l^{k-1}_n})\\
=&u^{k+1}(\frac{u^{k-1}_1}{2},\cdots,\frac{u^{k-1}_n}{2})\\
=&\frac{1}{2}u^{k+1}({u^{k-1}_1},\cdots,{u^{k-1}_n})\\
=&\frac{1}{2}u^{k+1}({u^{k}_1},\cdots,{u^{k}_n})\\
\geq &\frac{1}{2}max\{u^k_1,\cdots,u^k_n\}\\
=&\frac{1}{2}u_M^{k+1}({u^{k}_1},\cdots,{u^{k}_n})
\end{aligned}
$$

This means $u_M^{k+1}(\cdot)$ is the block-wise tightest.
\end{proof}

\end{document}